\documentclass{article}
\usepackage{iclr2026_conference,times}
\iclrfinalcopy
\usepackage[T1]{fontenc}
\usepackage[utf8]{inputenc}
\usepackage{booktabs,graphicx,amsmath,amssymb,array,tabularx}
\usepackage{placeins,flafter,float,algorithm}
\usepackage{xurl}
\usepackage[hidelinks]{hyperref}
\usepackage{microtype}
\usepackage{enumitem}
\usepackage{caption}
\setlist{nosep,leftmargin=*}
\title{VbotNav-0: Asynchronous Cloud--Edge Dual-Policy Navigation across Indoor and Outdoor Environments}\author{
Shiqi Tan\thanks{Equal contribution. Shiqi Tan and Yujun Zhang are co-first authors.} \\
{\small\href{mailto:shiqi.tan@vbot.cn}{\texttt{shiqi.tan@vbot.cn}}}
\And
Yujun Zhang\footnotemark[1] \\
{\small\href{mailto:yujun.zhang@vbot.cn}{\texttt{yujun.zhang@vbot.cn}}}
\And
Keyu Li \\
{\small\href{mailto:keyu.li@vbot.cn}{\texttt{keyu.li@vbot.cn}}}
\And
Shuaixi Pan \\
{\small\href{mailto:shuaixi.pan@vbot.cn}{\texttt{shuaixi.pan@vbot.cn}}}
\AND
Siyu Zhou\thanks{Corresponding authors: Siyu Zhou, Hailong Qin, and Yinan Yu.} \\
{\small\href{mailto:siyu.zhou@vbot.cn}{\texttt{siyu.zhou@vbot.cn}}}
\And
Hailong Qin\footnotemark[2] \\
{\small\href{mailto:hailong.qin@vbot.cn}{\texttt{hailong.qin@vbot.cn}}}
\And
Yinan Yu\footnotemark[2] \\
{\small\href{mailto:yinan.yu@vbot.cn}{\texttt{yinan.yu@vbot.cn}}}
}
\begin{document}
\maketitle
\fancyhead{} 
\renewcommand{\headrulewidth}{0pt}
\begin{abstract}
Continuous indoor--outdoor navigation connects destination-level reasoning with motion decisions that must respond to current observations. We present VbotNav-0, a cloud--edge dual-policy framework that links these processes through asynchronous policy-to-policy conditioning. The cloud policy produces subgoal guidance from visual history, navigation instructions, and route tokens; the edge policy combines available guidance with current RGB and LiDAR observations to generate local trajectories. This formulation allows the local perception--action loop to continue between cloud updates. A common route representation connects indoor paths and outdoor map routes, and the training recipe combines navigation demonstrations, corrective data, and grounded visual question answering (VQA). In the reported closed-loop evaluation of 258 episodes across 145 scenes, VbotNav-0 achieves 55.04\% success rate and 55.03\% success weighted by path length. 
\end{abstract}
\section{Introduction}
\label{sec:introduction}
A robot navigating from an outdoor road through a building entrance to an indoor destination must coordinate two decision horizons. Route-level decisions combine the supplied route with visual history and navigation instructions, whereas local motion depends on changing obstacles, traversability, and pose. Long-range reasoning is useful only if it can guide a physical feedback loop that remains responsive while the robot moves.

Cloud--edge deployment makes this coordination explicit. A large multimodal model can process visual history, instructions, and route context in the cloud, while an onboard policy generates local trajectories. Requiring every local decision to wait for cloud inference couples motion updates to computation and communication delay. Conversely, replaying cloud-generated motion without interpreting current observations limits adaptation to the immediate scene. The design question is how to preserve destination-level intent while allowing local navigation decisions to update independently.

We present \emph{VbotNav-0}, an asynchronous cloud--edge dual-policy framework. A cloud navigation policy generates subgoal guidance, and an edge navigation policy conditions local trajectory generation on that guidance and fresh onboard observations. Both components generate navigation decisions at their respective horizons; a separate locomotion controller executes the edge output. Between cloud arrivals, the edge reuses the available guidance while recomputing its trajectory. A new cloud result changes the condition for subsequent local decisions. We call this dependence \emph{policy-to-policy conditioning}.

Indoor planner paths and outdoor map routes are represented through a common route interface. Lookahead route tokens describe the upcoming route segment, allowing the cloud policy to relate the supplied route to visual observations and generate subgoal guidance.

Hierarchical and asynchronous navigation already connect slow reasoning to faster action generation~\citep{ChengA-RSS-25,gong2026abot,huang2026ticvla,yang2026asyncshield}. Our focus is a common route-conditioned interface between two navigation policies for indoor--outdoor operation, together with an explicit account of what guidance is available at each edge decision. This formulation separates the communication schedule from the local trajectory update and provides a basis for evaluating the effect of cloud-update delays.

The current empirical study evaluates overall navigation and its training components. On 258 episodes across 145 scenes, the reported main evaluation reaches 55.04\% SR and 55.03\% SPL. Separate ablations examine corrective navigation data and VQA co-training, while four VQA tasks assess semantic recognition, spatial localization, road rules, and waypoint diagnosis. We make three contributions:
\begin{enumerate}
\item A dual-policy formulation connecting cloud-generated subgoal guidance to onboard trajectory generation through a shared indoor--outdoor route interface.
\item An explicit asynchronous conditioning model that distinguishes context acquisition, result arrival, and local decisions, allowing trajectory updates between cloud results.
\item An empirical study of navigation, training components, and grounded understanding, with an explicit account of the evaluation scope.
\end{enumerate}

\section{Related Work}
\label{sec:related_work}
\paragraph{Hierarchical and asynchronous navigation.}
Generalist navigation models such as NavFoM, LightNav-0, and SPAN-Nav explore shared representations and spatial interfaces~\citep{zhang2025navfom,lightorigins2026lightnav,liu2026span}. NaVILA connects high-level spatial commands to a locomotion policy~\citep{ChengA-RSS-25}. More directly, ABot-N1 separates slow reasoning from fast waypoint generation under cached guidance~\citep{gong2026abot}; TIC-VLA incorporates delayed semantic states and latency information~\citep{huang2026ticvla}; and AsyncShield aligns cloud guidance and adapts it with an edge policy~\citep{yang2026asyncshield}. CloudEdgeVLA studies delayed cloud--edge conditioning for manipulation~\citep{peng2026cloudedgevla}. These works make asynchronous high-level guidance an established design direction. VbotNav-0 examines this direction through a common route interface and two navigation policies intended to span indoor and outdoor operation.

\paragraph{Route guidance and navigation settings.}
Room-to-Room and VLN-CE study language-guided navigation in discrete and continuous indoor environments~\citep{anderson2018vln,krantz2020continuous}. Semantic exploration supports object-goal navigation~\citep{chaplot2020semexp}, while ViNT learns goal-reaching behavior from heterogeneous robot data and NaVid predicts actions from video and language~\citep{pmlr-v229-shah23a,Zhang-RSS-24}. In outdoor settings, TOUCHDOWN studies spatial language in street imagery~\citep{chen2019touchdown}; CityWalker learns from web videos, UrbanVLA incorporates route guidance, and SocialNav connects social reasoning to trajectory generation~\citep{Liu_2025_CVPR,li2025urbanvla,chen2025socialnav}. NavVerse and out-to-in navigation address environment transitions~\citep{wu2026navverse,zhao2026outtoin}. Our task uses a supplied reference route, navigation instructions, and visual observations. The relevant distinction is the information available to each policy, rather than the geographic scale of its pretraining corpus.

\paragraph{Corrective and grounded supervision.}
DAgger addresses distribution shift by obtaining expert labels at policy-visited states~\citep{ross2011dagger}. Grounded supervision complements navigation imitation by associating semantic answers with spatial evidence; hallucination evaluation such as POPE highlights the need to check unsupported visual claims~\citep{li2023pope}. VbotNav-0 combines corrective navigation data with VQA tasks about traversability, localization, road rules, and proposed waypoints. We evaluate navigation and understanding separately so that gains in answer quality are not interpreted as direct measurements of physical safety.

\section{Problem Formulation}
\label{sec:problem_formulation}
\paragraph{Task and available information.}
We consider navigation from an initial pose to a destination along a supplied reference route, potentially crossing outdoor regions, entrances, and indoor spaces. The cloud policy receives a navigation instruction $\ell$, visual observations $\mathcal H_t$ comprising the current RGB image and historical frames, and route tokens $\mathcal G_t$. At local decision time $t$, $o_t=(I_t,L_t)$ denotes current RGB and LiDAR observations. A route window $\mathcal G_t=\{g_{t,i}\}_{i=1}^{K}$ contains lookahead geometry in the robot frame, together with source, availability, and reliability information. It is derived from the supplied route, not from future observations or executed ground-truth motion.

\paragraph{Policy outputs.}
The cloud policy $\pi^C_\theta$ maps a context snapshot acquired at time $\sigma_j$ to guidance $z_j$. The waypoint guidance in Figure~\ref{fig:unified-model} represents this local subgoal. The edge policy $\pi^E_\phi$ maps current observations and available guidance $\bar z_t$ to a trajectory:
\begin{equation}
\begin{aligned}
z_j &\sim \pi^C_\theta(\cdot\mid\mathcal H_{\sigma_j},\ell,\mathcal G_{\sigma_j}),\\
\widehat{\mathcal W}_t &\sim \pi^E_\phi(\cdot\mid o_t,\bar z_t).
\end{aligned}
\label{eq:model}
\end{equation}
The trajectory contains ten points, each with components $(x,y,\sin\psi,\cos\psi)$, and is executed by a separate locomotion controller. Policy-to-policy conditioning refers to conditioning on a navigation decision, not transmitting network weights.

\paragraph{Asynchronous availability.}
Let $a_j\geq\sigma_j$ be the arrival time of cloud result $j$. Among results available at $t$, the formulation selects the one with the newest source context:
\begin{equation}
j^*(t)=\operatorname*{arg\,max}_{j:a_j\leq t}\sigma_j,
\qquad \bar z_t=\mathcal A_t(z_{j^*(t)}).
\label{eq:async}
\end{equation}
$\mathcal A_t$ denotes the interpretation of guidance in the edge policy's representation and coordinate frame. Between cloud arrivals, the selected guidance can remain fixed while observations and trajectories change. Equations~\ref{eq:model}--\ref{eq:async} apply when usable guidance exists; startup, expiration, and communication loss require supervisory handling. Navigation ends at the destination or an evaluator-defined termination condition.

\section{Asynchronous Cloud--Edge Dual-Policy Navigation}
\label{sec:unified-model}
\begin{figure}[htbp]
\centering
\includegraphics[width=\linewidth,height=0.78\textheight,keepaspectratio]{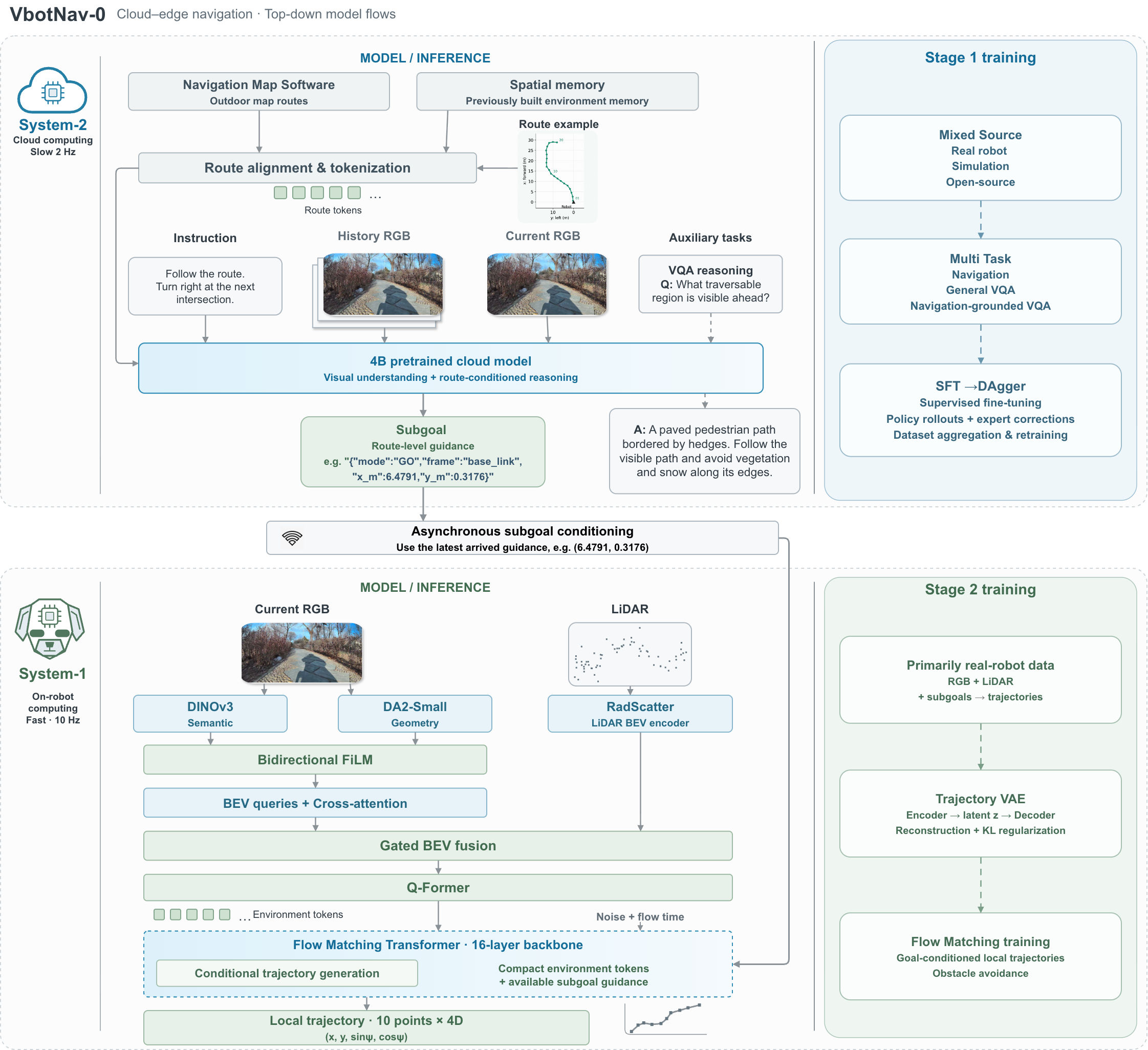}
\caption{VbotNav-0 architecture and training. Upper left: System 2 aligns and tokenizes routes from navigation map software or spatial memory, and combines them with instructions and RGB history in a 4B cloud model to predict subgoals; the auxiliary VQA branch illustrates question--answer supervision. Lower left: System 1 fuses RGB semantics and geometry with LiDAR BEV features, compresses them with a Q-Former, and generates ten-point trajectories via a subgoal-conditioned flow-matching Transformer. The policies communicate asynchronously at nominal rates of 2 and 10\,Hz. Right: Stage 1 combines mixed-source, multi-task SFT with DAgger; Stage 2 uses primarily real-robot data for trajectory-VAE learning and flow-matching-based local obstacle avoidance.}
\label{fig:unified-model}
\end{figure}

Figure~\ref{fig:unified-model} summarizes the cloud--edge architecture (left) and its two-stage training recipe (right). The cloud policy (System 2) interprets visual history, instructions, and route tokens to produce route-level subgoals. The onboard policy (System 1) combines the available subgoal with fresh RGB and LiDAR observations to generate local obstacle-avoiding trajectories. These policies operate asynchronously, as detailed in Section~\ref{sec:async-execution}; Stage 1 and Stage 2 denote training phases, not sequential inference steps. A downstream locomotion controller (System 0, not shown) converts trajectories into robot commands.

\subsection{Shared Route Interface}
Routes originate from navigation map software for outdoor travel or previously built spatial memory for known environments. Indoor planner paths derived from this memory and outdoor map routes are aligned and sampled into $K$ lookahead tokens. Expressing geometry in a local robot frame gives both sources a common connection to immediate motion. Source tags distinguish their origin; availability indicates whether a cue exists, and reliability characterizes an available cue. Progress identifies the relevant part of the route. Optional GPS can support alignment without changing the route interface.

The route window $\mathcal G_t$ identifies the upcoming segment of the supplied route. The cloud policy interprets this segment together with visual history and the navigation instruction to generate a local subgoal. This shared route representation connects indoor and outdoor route sources to the same cloud--edge policy interface.

\subsection{Cloud and Edge Navigation Policies}
The cloud policy uses a 4B pretrained model for scene understanding and longer-horizon route reasoning. Its inputs comprise the current RGB image, historical RGB frames, a navigation instruction, and route tokens. The predicted waypoint serves as a local subgoal for the edge policy. The auxiliary VQA branch provides supervision for traversability, road rules, region localization, and waypoint diagnosis; its understanding performance is evaluated separately from navigation.

The edge navigation model predicts local trajectories for obstacle avoidance under the available subgoal guidance. Current RGB observations are encoded by DINOv3 for semantics and DA2-Small for geometry. Bidirectional FiLM exchanges information between these features, and learnable BEV queries aggregate them through cross-attention. A RadScatter encoder separately maps LiDAR observations to BEV features. Gated fusion combines the camera and LiDAR representations, and a Q-Former compresses the fused features into environment tokens.

A 16-layer flow-matching Transformer conditions trajectory generation on these environment tokens and the available cloud subgoal, together with noise and flow time. Each output trajectory contains ten points with planar positions and headings encoded as $(x,y,\sin\psi,\cos\psi)$. System 0 converts the trajectory into low-level locomotion commands. The nominal update rates are 2\,Hz for cloud reasoning, 10\,Hz for edge prediction, and 50\,Hz for locomotion control. Between cloud updates, the edge policy recomputes trajectories from fresh observations using the latest available guidance.

\subsection{Asynchronous Conditioning and Execution}
\label{sec:async-execution}
\begin{figure}[htbp]
\centering
\includegraphics[width=\linewidth]{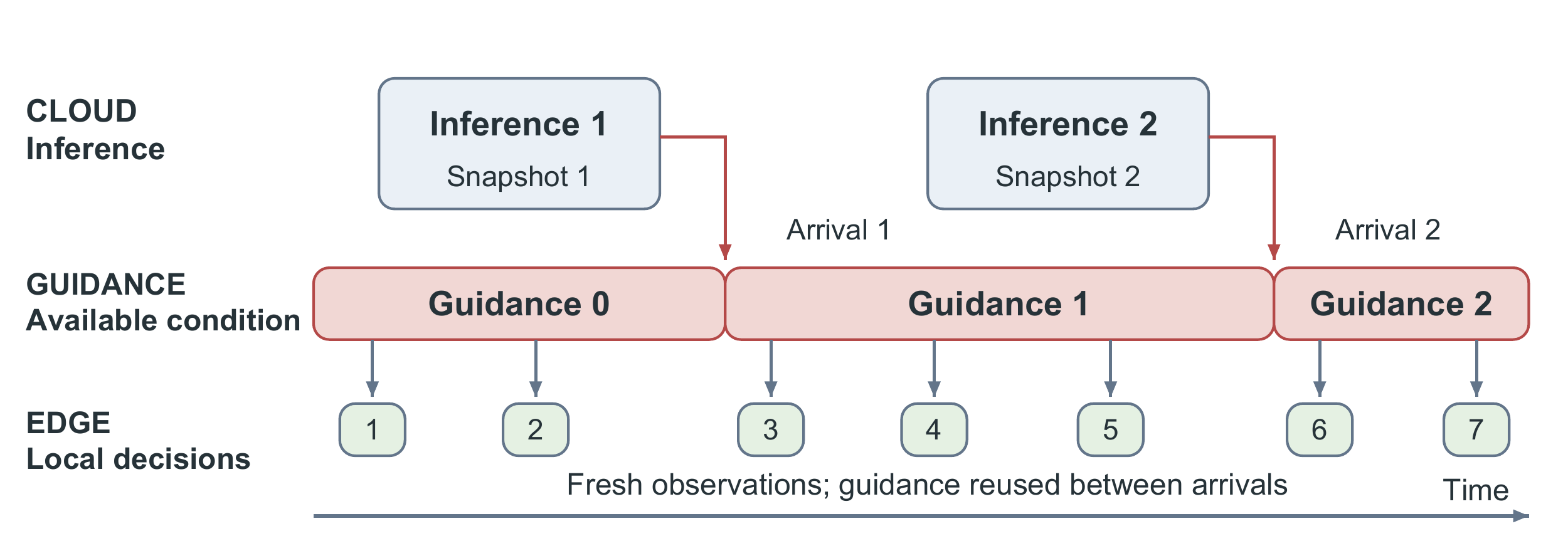}
\caption{Schematic asynchronous schedule, initialized with usable guidance, not a latency measurement. A context snapshot is processed in the cloud and becomes usable only after arrival. Multiple edge decisions reuse the available guidance while incorporating fresh observations. A newer result updates subsequent decisions; an in-flight result cannot condition an earlier decision.}
\label{fig:async_timeline}
\end{figure}

A cloud result updates the condition available to the edge; it does not prescribe every intervening trajectory. The edge therefore responds to fresh observations while retaining route-level intent. Figure~\ref{fig:async_timeline} distinguishes context acquisition, result arrival, and local decisions. Asynchrony is this nonblocking information dependency, rather than merely two nominal update frequencies.

\begin{algorithm}[htbp]
\caption{Interface-level inference schedule corresponding to Eq.~\ref{eq:async}. The schedule specifies dependencies, not a claim that timestamp ordering, guidance conversion, or fallback has been independently verified in deployment.}
\label{alg:async_schedule}
\centering
\small
\begin{tabularx}{\linewidth}{@{}rX@{}}
\toprule
1 & Independently acquire cloud context at $\sigma_j$ and generate guidance $z_j$. \\
2 & Make result $j$ eligible for edge use only after its arrival $a_j$. \\
3 & At each edge decision $t$, acquire fresh observations $o_t$. \\
4 & If usable guidance exists, select the newest available source context and interpret it using $\mathcal A_t$. \\
5 & Generate $\widehat{\mathcal W}_t$ conditioned on $o_t$ and $\bar z_t$; pass it to the locomotion controller. \\
6 & Otherwise, defer to the system's supervisory rule for missing or invalid guidance. \\
\bottomrule
\end{tabularx}
\end{algorithm}

The schedule requires guidance to remain interpretable as the robot moves. Whether this involves frame conversion, re-anchoring, or a learned condition depends on the actual output representation. The current formulation leaves this choice explicit. Likewise, an out-of-order arrival must not be confused with a newer context; Eq.~\ref{eq:async} orders by source time, and the transport implementation must be checked against that rule.

\subsection{Continuous Operation and Supervision}
The same interface is intended to span outdoor routes, entrances, and indoor paths. Supervision manages route progress and guidance validity around the two policies. The proposed state includes route segment, trusted subgoal, failure counts, and visual anchors, supporting decisions such as waiting, relocalization, replanning, and stopping. This supervisory design is separate from the learned trajectory output; recovery effectiveness requires episode-level evidence. The update rates in Figure~\ref{fig:unified-model} specify the configuration; achieved rates and communication latency require timing measurements.

\FloatBarrier
\section{Data and Training}
\label{sec:data}
\label{sec:training}
The training recipe follows the two stages in Figure~\ref{fig:unified-model}. We distinguish training sources from task-specific evaluation sets and describe supervision separately from its measured effects.

\paragraph{Stage 1: mixed-source, multi-task cloud-policy training.}
Real-robot, simulation, and open-source data support joint learning of navigation, general VQA, and navigation-grounded VQA. Supervised fine-tuning (SFT) initializes the cloud policy; subsequent DAgger-style iterations collect policy rollouts, obtain expert corrections, aggregate these examples with the training set, and retrain. The auxiliary question--answer path in the figure illustrates the visual-understanding supervision used alongside subgoal prediction.

\paragraph{Stage 2: local trajectory and obstacle-avoidance training.}
Training uses primarily real-robot examples pairing RGB and LiDAR observations and subgoal guidance with local trajectories. A trajectory VAE learns an encoder--latent--decoder representation through reconstruction and KL regularization. Flow-matching training learns goal-conditioned local trajectory generation for obstacle avoidance. This stage complements route-level cloud reasoning with an onboard policy that reacts to current geometry and obstacles.

\subsection{Navigation Corpus}
The reported corpus contains over 650,000 clips: 350,000+ real-world clips and 300,000+ simulation clips. The environment totals are 400,000+ outdoor and 250,000+ indoor clips.

\begin{figure}[htbp]
\centering
\includegraphics[width=\linewidth]{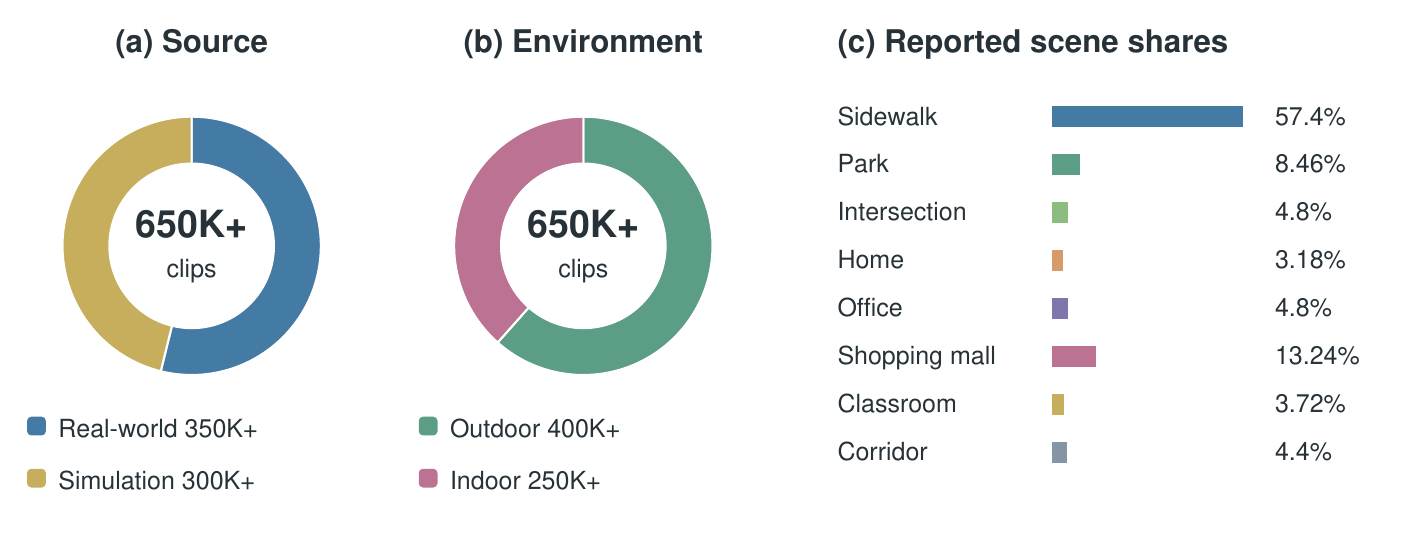}
\caption{Reported navigation-data composition. Source and environment panels use nominal counts while retaining the supplied ``+'' labels. The scene shares are shown separately.}
\label{fig:data_distribution}
\end{figure}

The separately reported scene shares are sidewalk 57.4\%, park 8.46\%, intersection 4.8\%, home 3.18\%, office 4.8\%, shopping mall 13.24\%, classroom 3.72\%, and corridor 4.4\%. The total duration of the collected data exceeds 1,000 hours.

\subsection{Navigation Imitation and Corrective Data}
Expert navigation trajectories provide the initialization signal. The described sources include indoor planners, outdoor navigation stacks, and human correction. DAgger-style aggregation~\citep{ross2011dagger} adds supervision at states visited by the current policy, including route deviations, oscillations, incorrect turns, and overshoot. Expert labels describe local waypoints and, where available, anomalies and recovery actions. The proposed aggregation protocol reduces expert mixing over successive rounds and includes event-balanced recovery sampling.

\subsection{Grounded VQA Supervision}
\label{sec:vqa}
VQA supplies auxiliary supervision for four navigation-relevant capabilities: road-rule understanding, traversable-class recognition, region localization, and waypoint diagnosis. Answers are associated with spatial evidence such as image regions or route geometry. The intended verification protocol uses available depth, occupancy, maps, OCR, detections, and odometry to check supported claims and retain uncertain cases as uncertain.
\FloatBarrier
\section{Experiments}
\label{sec:experiments}
We ask three empirical questions: how the complete evaluation pipeline compares with navigation baselines; how corrective data and VQA co-training affect navigation; and how VQA training changes grounded understanding. 

\subsection{Evaluation Setting and Protocol}
\label{sec:navigation-evaluation}
\paragraph{Navigation setting.}
The main evaluation contains 258 episodes across 145 self-collected scenes. Methods share saved routes, rendered scenes, and closed-loop termination rules. This describes the reported evaluator; the table should not be read as 258 verified physical-robot deployments.

\paragraph{Metrics.}
Success rate (SR) measures successful goal completion. Success weighted by path length (SPL) also penalizes excess path length on successful episodes. Both are reported as percentages, with higher values better. The non-traversable failure rate records episodes whose predicted endpoint lies in an annotated non-traversable region. Timeout includes exhausting a route-length-dependent step budget or making less than 0.25\,m progress along the reference route for three consecutive steps. Lower failure rates are better. All four main-table metrics use the same 258 episodes. 

\paragraph{Baselines and information access.}
We compare with CityWalker~\citep{Liu_2025_CVPR}, ViNT~\citep{pmlr-v229-shah23a}, NaVid~\citep{Zhang-RSS-24}, NaVILA~\citep{ChengA-RSS-25}, and SocialNav~\citep{chen2025socialnav}. Each is adapted to the common evaluator without retraining. CityWalker and SocialNav receive five causal RGB observations and a route-derived target; NaVid and NaVILA use route-derived language and their native history formatting; ViNT additionally receives a rendered goal-view preview. Thus, the environment and stopping rules are shared, but native information interfaces are not identical. Appendix~\ref{app:baseline} specifies the reported conversions.

\subsection{Navigation Results}

\begin{table}[tb]
\centering
\caption{Closed-loop navigation results on 258 episodes across 145 self-collected scenes. All values are percentages; best results in each column are bold.}
\label{tab:navigation_comparison}
\small
\renewcommand{\arraystretch}{1.12}
\begin{tabular*}{\linewidth}{@{\extracolsep{\fill}}lcccc@{}}
\toprule
\textbf{Method} &
\textbf{SR} $\uparrow$ &
\textbf{SPL} $\uparrow$ &
\textbf{Non-trav.} $\downarrow$ &
\textbf{Timeout} $\downarrow$ \\
\midrule
CityWalker & 17.83 & 17.83 & 59.69 & 15.89 \\
ViNT & 15.12 & 15.12 & 58.53 & \textbf{0.39} \\
NaVid-VLN-CE & 8.53 & 8.53 & \textbf{20.54} & 69.77 \\
NaVILA & 9.30 & 9.30 & 25.58 & 62.79 \\
SocialNav & 17.83 & 17.82 & 70.16 & 1.55 \\
\midrule
\textbf{VbotNav-0 (Ours)} &
\textbf{55.04} & \textbf{55.03} & 38.76 & 3.49 \\
\bottomrule
\end{tabular*}
\end{table}
\paragraph{Overall comparison.}
Table~\ref{tab:navigation_comparison} reports the highest SR (55.04\%) and SPL (55.03\%) for VbotNav-0. Relative to CityWalker, the strongest baseline by SPL, the gains are 37.21 and 37.20 percentage points, respectively. CityWalker and SocialNav share the highest baseline SR at 17.83\%.
VbotNav-0 has lower non-traversable failure rates than CityWalker and SocialNav, but does not minimize every failure category. NaVid-VLN-CE has the lowest non-traversable rate and a high timeout rate; ViNT has the lowest timeout rate but lower success.

\subsection{Training Ablations}
\label{sec:ablation}
Table~\ref{tab:navigation_ablation} compares the full model with variants removing DAgger data or VQA co-training.

\begin{table}[tb] \centering \caption{Ablation of DAgger data and VQA co-training. Each ablated variant removes the indicated component from VbotNav-0. Both metrics are percentages; higher is better. Best results are bold.} \label{tab:navigation_ablation} \renewcommand{\arraystretch}{1.12} \begin{tabular*}{\linewidth}{@{\extracolsep{\fill}}lcc@{}} \toprule \textbf{Model variant} & \textbf{SR (\%)} $\uparrow$ & \textbf{SPL (\%)} $\uparrow$ \\ \midrule \textbf{VbotNav-0 (Ours)} & \textbf{55.04} & \textbf{55.03} \\ w/o DAgger data & 38.03 & 37.98 \\ w/o VQA co-training & 45.81 & 45.68 \\ \bottomrule \end{tabular*}
\end{table}
Removing DAgger data reduces SR by 17.01 percentage points and SPL by 17.05 points. Removing VQA co-training reduces SR by 9.23 points and SPL by 9.35 points. The results are consistent with benefits from corrective and semantic supervision, with the larger SR decrease observed for DAgger removal.

\subsection{Grounded VQA Evaluation}
\label{sec:vqa-evaluation}
\paragraph{Tasks and metrics.}
We compare the base model and the VQA-trained model on four tasks. Traversable-class recognition uses micro-F1 after category normalization and deduplication. Region localization uses class-consistent, one-to-one greedy box matching at IoU $\geq0.50$, followed by micro-F1. Road rules and waypoint diagnosis use a fixed GLM-4.6V-Flash judge with a 0--4 semantic rubric; scores of at least 3 pass. Appendix~\ref{app:vqa-metrics} gives the matching rules, equations, numerical tolerance, and scoring rubric.

\begin{table}[tb]
\centering
\caption{Effect of VQA training on semantic understanding and spatial grounding. All values are percentages ($\uparrow$); best results are bold.}
\label{tab:vqa_training}
\vspace{3pt}
\small
\setlength{\tabcolsep}{3pt}
\renewcommand{\arraystretch}{1.2}
\begin{tabular*}{\linewidth}{@{\extracolsep{\fill}}ccccc@{}}
\toprule
\textbf{Model variant} & \shortstack{\textbf{Road rules}\\Pass rate} & \shortstack{\textbf{Traversable}\\\textbf{classes}\\F1} & \shortstack{\textbf{Region}\\\textbf{localization}\\F1@0.50} & \shortstack{\textbf{Waypoint}\\\textbf{diagnosis}\\Pass rate} \\
\midrule
Base (w/o VQA) & 33.0 & 40.5 & 12.4 & 3.5 \\
\midrule
\textbf{VQA-trained} & \textbf{59.3} & \textbf{87.2} & \textbf{59.1} & \textbf{45.0} \\
\bottomrule
\end{tabular*}
\end{table}
\paragraph{Results.}
VQA training improves all four measures in Table~\ref{tab:vqa_training}: road-rule pass rate by 26.3 percentage points, traversable-class F1 by 46.7, localization F1 by 46.7, and waypoint-diagnosis pass rate by 41.5. The gains support improved semantic recognition, grounding, and route-related reasoning. The remaining 45.0\% waypoint-diagnosis pass rate also indicates substantial residual error. Semantic pass rates and F1 measure different properties and are not a shared difficulty scale.

\paragraph{Qualitative evidence.}
Figures~\ref{fig:vqa_qualitative}--\ref{fig:vqa_rules} show improvements alongside residual mistakes. The trained model recovers a crosswalk and separates sidewalk regions, although one extra box remains in the crosswalk example. It identifies a 5.81\,m route offset against a 5.82\,m reference and recognizes a valid route that the base model rejects. Road-rule responses add an explicit stop at a crossing and replace an unsupported traffic-side assumption with yielding to vehicles. These selected examples illustrate the aggregate task improvements without implying error-free behavior.

\subsection{Qualitative Navigation and Failure Analysis}
\label{sec:qualitative}
\paragraph{Matched evaluation episodes.}
Figures~\ref{fig:matched_bridge}--\ref{fig:matched_stairs} in Appendix~\ref{app:cases} show all six methods at the same starting pose, current camera view, and visible route, with each method's native observation history. VbotNav-0 selects a traversable initial goal in the bridge, plaza, and stair examples. Some baselines select a non-traversable first endpoint; others produce an accepted initial point and fail later through stalling or subsequent non-traversable predictions. A valid first prediction therefore does not establish episode success. The supplied figures label VbotNav-0 as ``CivicNav''; this historical figure label denotes the same reported method.

\paragraph{Route-level consistency.}
Figure~\ref{fig:navigation_demo} pairs four egocentric predictions with global route progress. The model proceeds through a clear doorway, initiates a left turn when locally visible, follows a later right turn, and issues Stop at the endpoint. These snapshots allow inspection of consistency between local geometry, action text, and route position.

\section{Discussion and Limitations}
\label{sec:limitations}

\paragraph{Dependence on preplanned indoor routes.}
Our indoor navigation pipeline assumes that a reference route has been planned before execution and can be converted into route tokens. The cloud policy reasons along this supplied route, while the edge policy handles local trajectory generation and obstacle avoidance. Local adaptation therefore does not remove the dependence on prior route information. In an unmapped or first-visit indoor environment where no reference route is available, the current system cannot be deployed out of the box; mapping or route preparation is required before navigation. Extending the system to such settings would require online exploration, spatial mapping, and route generation, together with mechanisms for updating guidance as new areas are observed. These capabilities remain future work.

\section{Conclusion}
\label{sec:conclusion}
VbotNav-0 formulates cloud--edge navigation as two policies coupled through asynchronous guidance. The cloud supplies subgoal intent, while the edge generates local trajectories from current observations under the available condition. The reported navigation comparisons, training ablations, and VQA evaluation provide evidence for task performance and the value of corrective and semantic supervision.

\FloatBarrier
\bibliographystyle{iclr2026_conference}
\bibliography{references}

\begin{thebibliography}{21}
\providecommand{\natexlab}[1]{#1}
\providecommand{\url}[1]{\texttt{#1}}
\expandafter\ifx\csname urlstyle\endcsname\relax
  \providecommand{\doi}[1]{doi: #1}\else
  \providecommand{\doi}{doi: \begingroup \urlstyle{rm}\Url}\fi

\bibitem[Anderson et~al.(2018)Anderson, Wu, Teney, Bruce, Johnson, S{\"u}nderhauf, Reid, Gould, and van~den Hengel]{anderson2018vln}
Peter Anderson, Qi~Wu, Damien Teney, Jake Bruce, Mark Johnson, Niko S{\"u}nderhauf, Ian Reid, Stephen Gould, and Anton van~den Hengel.
\newblock Vision-and-language navigation: Interpreting visually-grounded navigation instructions in real environments.
\newblock In \emph{Proceedings of the IEEE Conference on Computer Vision and Pattern Recognition}, pp.\  3674--3683, 2018.
\newblock URL \url{https://openaccess.thecvf.com/content_cvpr_2018/html/Anderson_Vision-and-Language_Navigation_Interpreting_CVPR_2018_paper.html}.

\bibitem[Chaplot et~al.(2020)Chaplot, Gandhi, Gupta, and Salakhutdinov]{chaplot2020semexp}
Devendra~Singh Chaplot, Dhiraj Gandhi, Abhinav Gupta, and Ruslan Salakhutdinov.
\newblock Object goal navigation using goal-oriented semantic exploration.
\newblock In \emph{Advances in Neural Information Processing Systems}, 2020.
\newblock URL \url{https://www.cs.cmu.edu/afs/cs/user/dchaplot/www/projects/semantic-exploration.html}.

\bibitem[Chen et~al.(2019)Chen, Suhr, Misra, Snavely, and Artzi]{chen2019touchdown}
Howard Chen, Alane Suhr, Dipendra Misra, Noah Snavely, and Yoav Artzi.
\newblock {TOUCHDOWN}: Natural language navigation and spatial reasoning in visual street environments.
\newblock In \emph{Proceedings of the IEEE/CVF Conference on Computer Vision and Pattern Recognition}, pp.\  12538--12547, 2019.
\newblock URL \url{https://openaccess.thecvf.com/content_CVPR_2019/html/Chen_TOUCHDOWN_Natural_Language_Navigation_and_Spatial_Reasoning_in_Visual_Street_CVPR_2019_paper.html}.

\bibitem[Chen et~al.(2025)Chen, Guo, Chu, Luo, Shen, Sun, Hu, Xie, Yang, Shi, Gu, Liu, Han, Wu, Xu, and Zhang]{chen2025socialnav}
Ziyi Chen, Yingnan Guo, Zedong Chu, Minghua Luo, Yanfen Shen, Mingchao Sun, Junjun Hu, Shichao Xie, Kuan Yang, Pei Shi, Zhining Gu, Lu~Liu, Honglin Han, Xiaolong Wu, Mu~Xu, and Yu~Zhang.
\newblock {SocialNav}: Training human-inspired foundation model for socially-aware embodied navigation.
\newblock \emph{arXiv preprint arXiv:2511.21135}, 2025.
\newblock URL \url{https://arxiv.org/abs/2511.21135}.

\bibitem[Cheng et~al.(2025)Cheng, Ji, Yang, Gongye, Zou, Kautz, Biyik, Yin, Liu, and Wang]{ChengA-RSS-25}
An-Chieh Cheng, Yandong Ji, Zhaojing Yang, Zaitian Gongye, Xueyan Zou, Jan Kautz, Erdem Biyik, Hongxu Yin, Sifei Liu, and Xiaolong Wang.
\newblock {NaVILA: Legged Robot Vision-Language-Action Model for Navigation}.
\newblock In \emph{Proceedings of Robotics: Science and Systems}, Los Angeles, CA, USA, June 2025.
\newblock \doi{10.15607/RSS.2025.XXI.018}.
\newblock URL \url{https://www.roboticsproceedings.org/rss21/p018.html}.

\bibitem[Gong et~al.(2026)Gong, Guo, Hu, Kong, Leng, Li, Li, Liu, Liu, Lu, Luo, Ming, Shen, Tao, Wang, Yin, Gu, Guan, Guo, Liu, Pang, Yang, Ye, Geng, Gu, Han, Jing, Pan, Sun, Yang, Zhang, Chen, He, Mei, Shi, Yang, Zhu, Cai, Ma, Su, Tang, Zheng, Chu, Wu, Tang, and Xu]{gong2026abot}
Ruiyan Gong, Yingnan Guo, Junjun Hu, Jintao Kong, Xiaoxu Leng, Tianlun Li, Weize Li, Fei Liu, Zhicheng Liu, Jia Lu, Minghua Luo, Chenlin Ming, Yanfen Shen, Jiyue Tao, Zhengbo Wang, Mingyang Yin, Minqi Gu, Zihao Guan, Wei Guo, Guoqing Liu, Huachong Pang, Menglin Yang, Zeqian Ye, Xiaoxiao Geng, Zhining Gu, Honglin Han, Di~Jing, Hongyu Pan, Mingchao Sun, Kuan Yang, Jianfang Zhang, Yanghong Chen, Ye~He, Wei Mei, Jiahao Shi, Xiangpo Yang, Yanqing Zhu, Yang Cai, Jingjing Ma, Shihui Su, Zixiao Tang, Linbo Zheng, Zedong Chu, Xiaolong Wu, Wenbin Tang, and Mu~Xu.
\newblock {ABot-N1}: Toward a general visual language navigation foundation model.
\newblock \emph{arXiv preprint arXiv:2607.10383}, 2026.
\newblock \doi{10.48550/arXiv.2607.10383}.
\newblock URL \url{https://arxiv.org/abs/2607.10383}.

\bibitem[Huang et~al.(2026)Huang, Zhang, Liu, Song, Tang, and Ma]{huang2026ticvla}
Zhiyu Huang, Yun Zhang, Johnson Liu, Rui Song, Chen Tang, and Jiaqi Ma.
\newblock {TIC-VLA}: A think-in-control vision-language-action model for robot navigation in dynamic environments.
\newblock In \emph{Proceedings of the International Conference on Machine Learning (ICML)}, 2026.
\newblock URL \url{https://ucla-mobility.github.io/TIC-VLA/}.

\bibitem[Krantz et~al.(2020)Krantz, Wijmans, Majumdar, Batra, and Lee]{krantz2020continuous}
Jacob Krantz, Erik Wijmans, Arjun Majumdar, Dhruv Batra, and Stefan Lee.
\newblock Beyond the nav-graph: Vision-and-language navigation in continuous environments.
\newblock In \emph{European Conference on Computer Vision}, 2020.
\newblock URL \url{https://arxiv.org/abs/2004.02857}.

\bibitem[Li et~al.(2025)Li, Wang, Zhang, Li, Qi, Chen, Zhang, and Wang]{li2025urbanvla}
Anqi Li, Zhiyong Wang, Jiazhao Zhang, Minghan Li, Yunpeng Qi, Zhibo Chen, Zhizheng Zhang, and He~Wang.
\newblock {UrbanVLA}: A vision-language-action model for urban micromobility.
\newblock \emph{arXiv preprint arXiv:2510.23576}, 2025.
\newblock \doi{10.48550/arXiv.2510.23576}.
\newblock URL \url{https://arxiv.org/abs/2510.23576}.

\bibitem[Li et~al.(2023)Li, Du, Zhou, Wang, Zhao, and Wen]{li2023pope}
Yifan Li, Yifan Du, Kun Zhou, Jinpeng Wang, Xin Zhao, and Ji-Rong Wen.
\newblock Evaluating object hallucination in large vision-language models.
\newblock In \emph{Proceedings of the 2023 Conference on Empirical Methods in Natural Language Processing}, pp.\  292--305. Association for Computational Linguistics, 2023.
\newblock \doi{10.18653/v1/2023.emnlp-main.20}.
\newblock URL \url{https://aclanthology.org/2023.emnlp-main.20/}.

\bibitem[{Light Origins Team}(2026)]{lightorigins2026lightnav}
{Light Origins Team}.
\newblock {LightNav-0}: Eliciting {VLM} spatial intelligence for generalist embodied navigation.
\newblock Official project repository, 2026.
\newblock URL \url{https://github.com/lightorigins/LightNav-0}.
\newblock Accessed September 8, 2026.

\bibitem[Liu et~al.(2026)Liu, Xu, Chen, Yue, Zhang, Wang, Li, Zhao, Li, Su, Zhang, and Wang]{liu2026span}
Jiahang Liu, Tianyu Xu, Jiawei Chen, Lu~Yue, Jiazhao Zhang, Zhiyong Wang, Minghan Li, Qisheng Zhao, Anqi Li, Qi~Su, Zhizheng Zhang, and He~Wang.
\newblock {SPAN-Nav}: Generalized spatial awareness for versatile vision-language navigation.
\newblock \emph{arXiv preprint arXiv:2603.09163}, 2026.
\newblock \doi{10.48550/arXiv.2603.09163}.
\newblock URL \url{https://arxiv.org/abs/2603.09163}.

\bibitem[Liu et~al.(2025)Liu, Li, Jiang, Sujay, Yang, Zhang, Abanes, Zhang, and Feng]{Liu_2025_CVPR}
Xinhao Liu, Jintong Li, Yicheng Jiang, Niranjan Sujay, Zhicheng Yang, Juexiao Zhang, John Abanes, Jing Zhang, and Chen Feng.
\newblock {CityWalker}: Learning embodied urban navigation from web-scale videos.
\newblock In \emph{Proceedings of the IEEE/CVF Conference on Computer Vision and Pattern Recognition (CVPR)}, pp.\  6875--6885, June 2025.
\newblock URL \url{https://openaccess.thecvf.com/content/CVPR2025/html/Liu_CityWalker_Learning_Embodied_Urban_Navigation_from_Web-Scale_Videos_CVPR_2025_paper.html}.

\bibitem[Peng et~al.(2026)Peng, Ma, Wang, Wang, and Ma]{peng2026cloudedgevla}
Daojie Peng, Fulong Ma, Bingtao Wang, Sheng Wang, and Jun Ma.
\newblock Latency-tolerant cloud-edge collaborative vision-language-action models via emergent representational specialization, 2026.
\newblock URL \url{https://arxiv.org/abs/2608.00569}.

\bibitem[Ross et~al.(2011)Ross, Gordon, and Bagnell]{ross2011dagger}
St{\'e}phane Ross, Geoffrey~J. Gordon, and J.~Andrew Bagnell.
\newblock A reduction of imitation learning and structured prediction to no-regret online learning.
\newblock In \emph{Proceedings of the Fourteenth International Conference on Artificial Intelligence and Statistics}, volume~15 of \emph{Proceedings of Machine Learning Research}, pp.\  627--635, 2011.
\newblock URL \url{https://proceedings.mlr.press/v15/ross11a.html}.

\bibitem[Shah et~al.(2023)Shah, Sridhar, Dashora, Stachowicz, Black, Hirose, and Levine]{pmlr-v229-shah23a}
Dhruv Shah, Ajay Sridhar, Nitish Dashora, Kyle Stachowicz, Kevin Black, Noriaki Hirose, and Sergey Levine.
\newblock {ViNT}: A foundation model for visual navigation.
\newblock In Jie Tan, Marc Toussaint, and Kourosh Darvish (eds.), \emph{Proceedings of The 7th Conference on Robot Learning}, volume 229 of \emph{Proceedings of Machine Learning Research}, pp.\  711--733. PMLR, 06--09 Nov 2023.
\newblock URL \url{https://proceedings.mlr.press/v229/shah23a.html}.

\bibitem[Wu et~al.(2026)Wu, Hu, Han, Chang, Dong, Rabiei, and Ghaffari]{wu2026navverse}
Junzhe Wu, Yue Hu, Zeyu Han, Po-Hsun Chang, Yinan Dong, Behrad Rabiei, and Maani Ghaffari.
\newblock {NavVerse}: Benchmarking indoor-to-outdoor embodied navigation in continuous robot simulation.
\newblock \emph{arXiv preprint arXiv:2607.19695}, 2026.
\newblock \doi{10.48550/arXiv.2607.19695}.
\newblock URL \url{https://arxiv.org/abs/2607.19695}.

\bibitem[Yang et~al.(2026)Yang, Chu, Guo, Wang, Xie, Shen, Wu, Li, and Xu]{yang2026asyncshield}
Kai Yang, Zedong Chu, Yingnan Guo, Zhengbo Wang, Shichao Xie, Yanfen Shen, Xiaolong Wu, Xing Li, and Mu~Xu.
\newblock {AsyncShield}: A plug-and-play edge adapter for asynchronous cloud-based {VLA} navigation, 2026.
\newblock URL \url{https://arxiv.org/abs/2604.24086}.

\bibitem[Zhang et~al.(2024)Zhang, Wang, Xu, Zhou, Hong, Fang, Wu, Zhang, and Wang]{Zhang-RSS-24}
Jiazhao Zhang, Kunyu Wang, Rongtao Xu, Gengze Zhou, Yicong Hong, Xiaomeng Fang, Qi~Wu, Zhizheng Zhang, and He~Wang.
\newblock {NaVid: Video-based VLM Plans the Next Step for Vision-and-Language Navigation}.
\newblock In \emph{Proceedings of Robotics: Science and Systems}, Delft, Netherlands, July 2024.
\newblock \doi{10.15607/RSS.2024.XX.079}.
\newblock URL \url{https://www.roboticsproceedings.org/rss20/p079.html}.

\bibitem[Zhang et~al.(2025)Zhang, Li, Qi, Li, Liu, Wang, Liu, Zhou, Wu, Li, Fan, Li, Chen, Gao, Wu, Zhang, and Wang]{zhang2025navfom}
Jiazhao Zhang, Anqi Li, Yunpeng Qi, Minghan Li, Jiahang Liu, Shaoan Wang, Haoran Liu, Gengze Zhou, Yuze Wu, Xingxing Li, Yuxin Fan, Wenjun Li, Zhibo Chen, Fei Gao, Qi~Wu, Zhizheng Zhang, and He~Wang.
\newblock Embodied navigation foundation model.
\newblock \emph{arXiv preprint arXiv:2509.12129}, 2025.
\newblock \doi{10.48550/arXiv.2509.12129}.
\newblock URL \url{https://arxiv.org/abs/2509.12129}.

\bibitem[Zhao et~al.(2026)Zhao, Yang, Zhu, Shen, Wang, Gu, Shi, Guo, and Xu]{zhao2026outtoin}
Yuxiang Zhao, Yirong Yang, Yanqing Zhu, Yanfen Shen, Chiyu Wang, Zhining Gu, Pei Shi, Wei Guo, and Mu~Xu.
\newblock Bridging the indoor-outdoor gap: Vision-centric instruction-guided embodied navigation for the last meters.
\newblock \emph{arXiv preprint arXiv:2602.06427}, 2026.
\newblock \doi{10.48550/arXiv.2602.06427}.
\newblock URL \url{https://arxiv.org/abs/2602.06427}.

\end{thebibliography}
\clearpage
\appendix
\section{Implementation and Evaluation Details}
\label{app:details}
This appendix records the available evaluation protocol and original qualitative evidence.

\subsection{Baseline Interfaces}
\label{app:baseline}
CityWalker and SocialNav receive five causal RGB observations and a target derived from the last point of the visible route. We transform the observed positions and target into the coordinate convention expected by each model, then convert the last waypoint in its predicted sequence back to a metric local goal. SocialNav retains its native navigation prompt and waypoint sampler. NaVid and NaVILA receive route-derived language instructions and image histories formatted for their respective native interfaces. We parse their predicted navigation actions and convert them into local goals. ViNT receives an observation history and a goal-view preview rendered from the visible route; we use the last waypoint in its predicted sequence as its local goal.

Native histories and ViNT's goal preview remain part of the inputs even when not displayed in the matched-episode figures.

\subsection{Grounded VQA Metrics}
\label{app:vqa-metrics}
\paragraph{Deterministic matching metrics.}
Traversable-Class F1 evaluates category recognition after mapping labels to a shared vocabulary (ground, ramp, road, sidewalk, stairs, and tactile paving) and removing duplicate categories per example. True positives (TP), false positives (FP), and false negatives (FN) are accumulated over the evaluation set. Region Localization F1 additionally requires correct spatial grounding: normalized boxes of the same class are matched greedily in descending intersection over union (IoU), with one-to-one assignment and an IoU threshold of 0.50. Unmatched or malformed predictions count as FP, and unmatched reference regions as FN. Because the evaluated outputs do not provide calibrated confidence scores, region localization is evaluated at one IoU threshold rather than with average precision. Both metrics use micro-averaged counts:
\[
\mathrm{F1}=100\,\frac{2\mathrm{TP}}{2\mathrm{TP}+\mathrm{FP}+\mathrm{FN}},
\qquad
\mathrm{IoU}(B_p,B_g)=\frac{|B_p\cap B_g|}{|B_p\cup B_g|}.
\]

\paragraph{Semantic pass rates.}
Road-rule understanding and waypoint diagnosis are scored by a fixed GLM-4.6V-Flash judge using a predefined 0--4 rubric, from incorrect to fully correct; scores of at least 3 (mostly correct) pass. For each task with $N$ examples,
\[
\mathrm{PassRate}=\frac{100}{N}\sum_{i=1}^{N}\mathbb{I}[s_i\geq3].
\]
For road rules, the judge receives the latest rectified camera image, reference answer, and prediction; the image is primary evidence for visible conditions. Scoring covers pedestrian-accessible areas, yielding, crossings, traffic controls, and unsafe surface transitions. For waypoint diagnosis, the reference answer is authoritative: defect type, direction, counts, waypoint indices, and displacement must be correct. Counts and indices require exact agreement, while displacement allows an absolute error of at most $0.10\,\mathrm{m}$. Both tasks penalize unsupported claims, contradictions, and critical omissions.

\paragraph{Ordinal rubric.}
Scores denote: 4, fully correct; 3, mostly correct with minor omissions; 2, partially correct with a major error; 1, mostly incorrect; and 0, incorrect or unusable. The pass threshold is 3. 

\clearpage
\subsection{Qualitative VQA Evidence}
\begin{figure}[H]
\centering
\includegraphics[width=0.80\linewidth,viewport=0 409 248 629,clip]{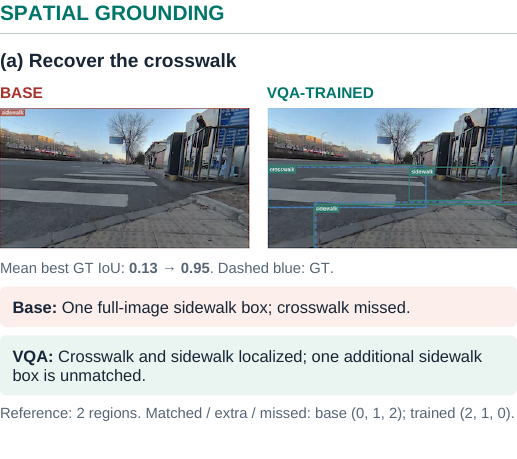}\\[8pt]
\includegraphics[width=0.80\linewidth,viewport=262 409 510 629,clip]{vqa_qualitative_evidence.pdf}
\caption{Spatial grounding before and after VQA training. Original boxes and IoU annotations are preserved. The crosswalk example retains an extra prediction.}
\label{fig:vqa_qualitative}
\end{figure}

\clearpage
\begin{figure}[H]
\centering
\includegraphics[width=0.80\linewidth,viewport=0 199 248 409,clip]{vqa_qualitative_evidence.pdf}\\[8pt]
\includegraphics[width=0.80\linewidth,viewport=262 199 510 409,clip]{vqa_qualitative_evidence.pdf}
\caption{Waypoint diagnosis before and after VQA training. The blue reference routes are shown only for evaluation.}
\label{fig:vqa_waypoints}
\end{figure}

\clearpage
\begin{figure}[H]
\centering
\includegraphics[width=0.80\linewidth,viewport=0 9 248 199,clip]{vqa_qualitative_evidence.pdf}\\[8pt]
\includegraphics[width=0.80\linewidth,viewport=262 9 510 199,clip]{vqa_qualitative_evidence.pdf}
\caption{Road-rule understanding before and after VQA training. Answers are condensed from the supplied examples; original observations are preserved.}
\label{fig:vqa_rules}
\end{figure}

\clearpage
\subsection{Matched Navigation Cases}
\label{app:cases}
The following figures retain the supplied observations, predictions, and endpoint annotations. ``CivicNav'' is the historical label for VbotNav-0 in these source images. Each figure compares a common starting situation, while later outcome annotations refer to the associated closed-loop rollout. They are selected cases rather than an estimate of failure frequency.

\begin{figure}[htbp]
\centering
\includegraphics[width=\linewidth]{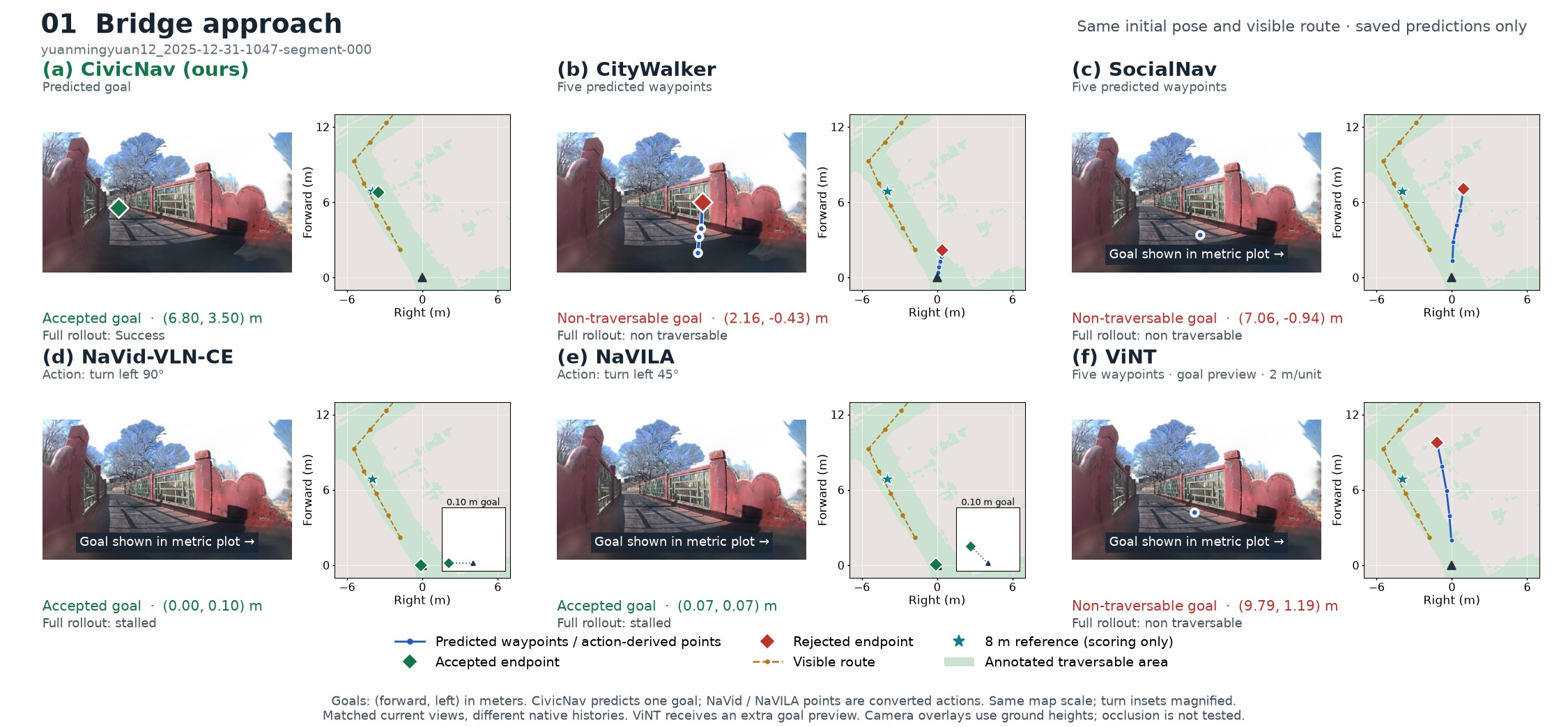}
\caption{First predictions on a bridge approach. VbotNav-0 selects a traversable goal along the bridge corridor. CityWalker, SocialNav, and ViNT select non-traversable endpoints; NaVid and NaVILA initially produce accepted turn-derived goals but later stall.}
\label{fig:matched_bridge}
\end{figure}

\begin{figure}[htbp]
\centering
\includegraphics[width=\linewidth]{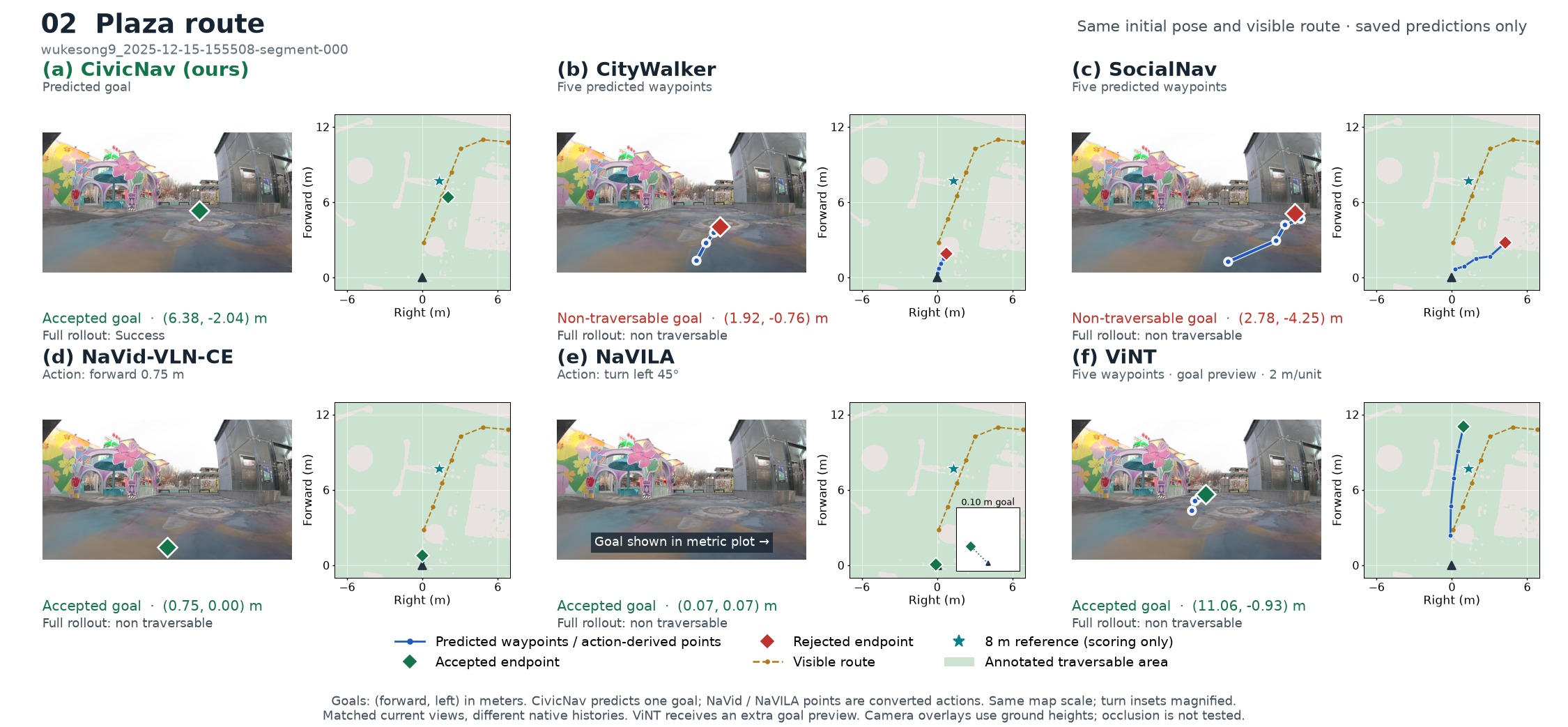}
\caption{First predictions on a plaza route. VbotNav-0 selects a traversable goal along the route and succeeds. CityWalker and SocialNav select non-traversable first endpoints. The other baselines have accepted first endpoints but fail later in their closed-loop rollouts.}
\label{fig:matched_plaza}
\end{figure}

\begin{figure}[htbp]
\centering
\includegraphics[width=\linewidth]{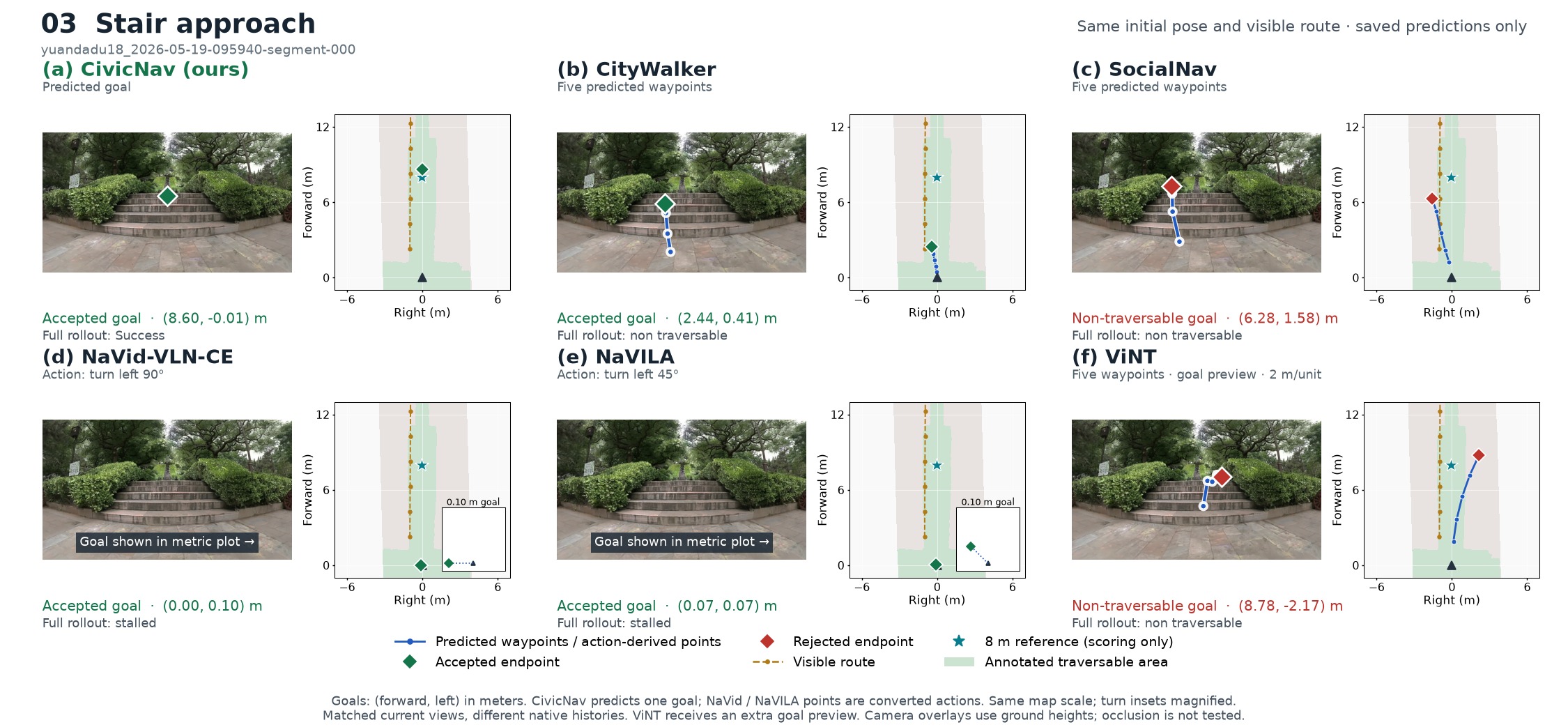}
\caption{First predictions on a stair approach. VbotNav-0 selects a goal within the annotated traversable corridor and succeeds. SocialNav and ViNT select non-traversable first endpoints. CityWalker initially selects an accepted endpoint but fails later, while NaVid and NaVILA subsequently stall.}
\label{fig:matched_stairs}
\end{figure}

\FloatBarrier
\subsection{Indoor--Outdoor Rollout Snapshots}
The snapshots below provide the original observation, route-progress view, and displayed rationale for four decisions. The rationale is model output; its presence does not verify that an action was safely executed. This distinction is important at crossings and around vehicles.
\begin{figure}[p]
    \centering
    \includegraphics[width=0.9\linewidth]{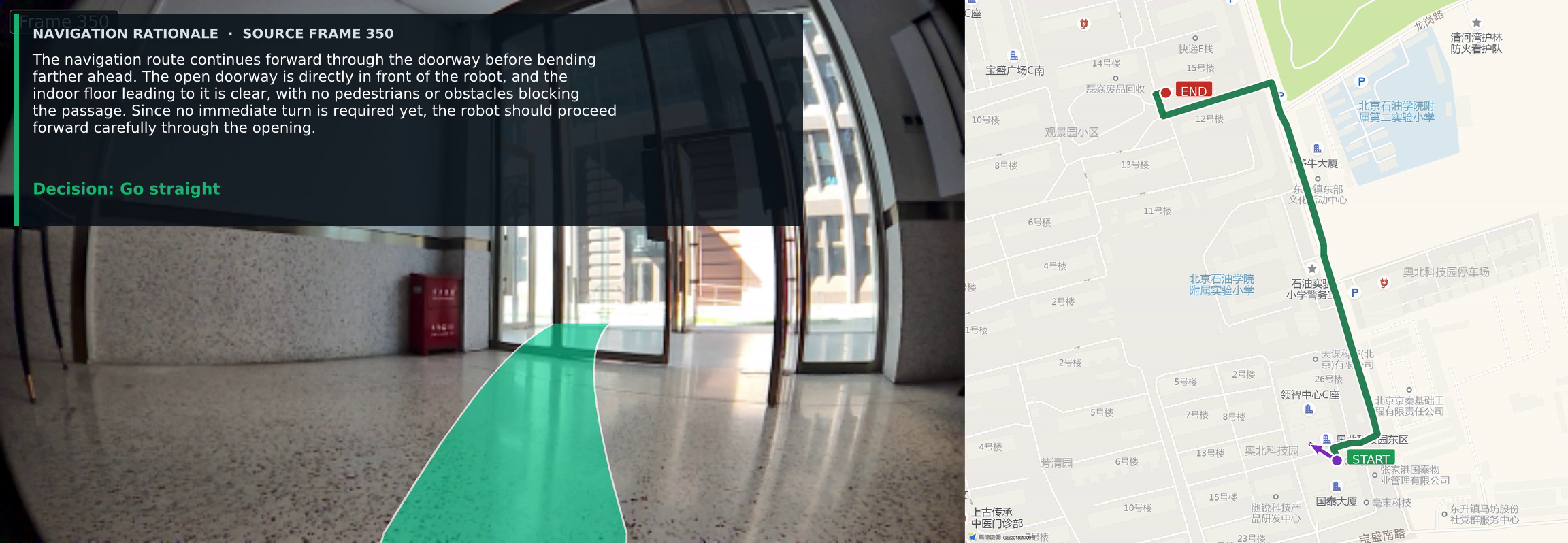}\\[-0.5mm] \includegraphics[width=0.9\linewidth]{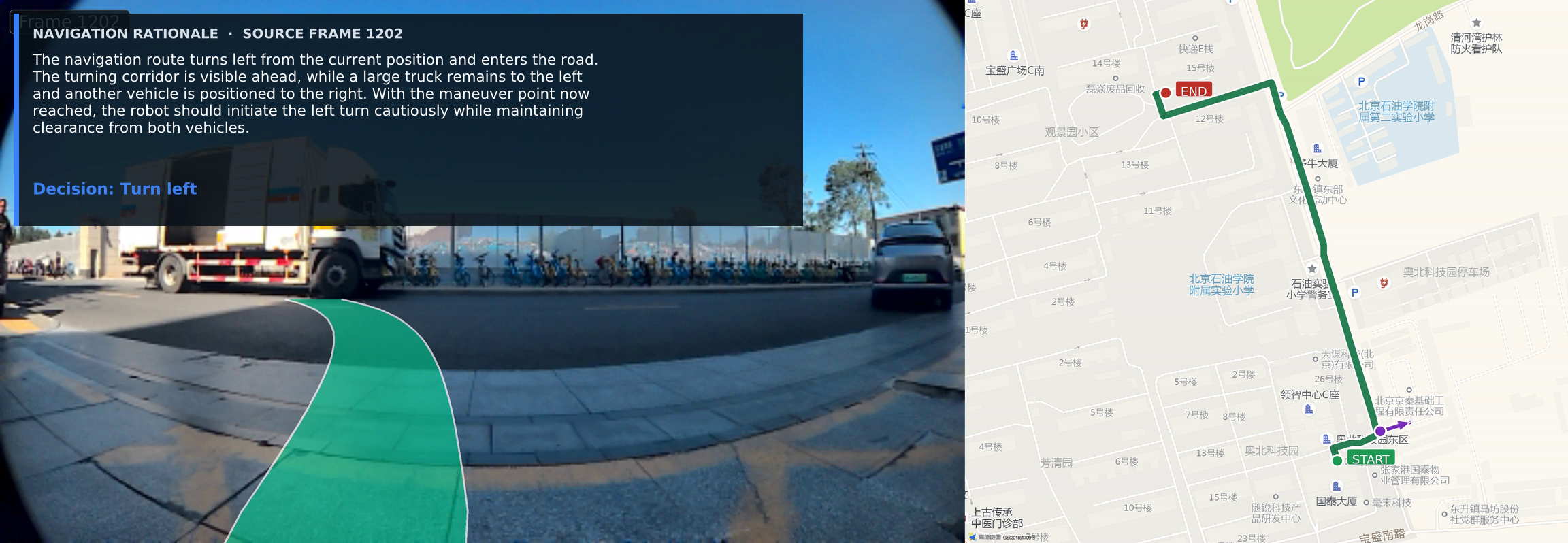}\\[-0.5mm] \includegraphics[width=0.9\linewidth]{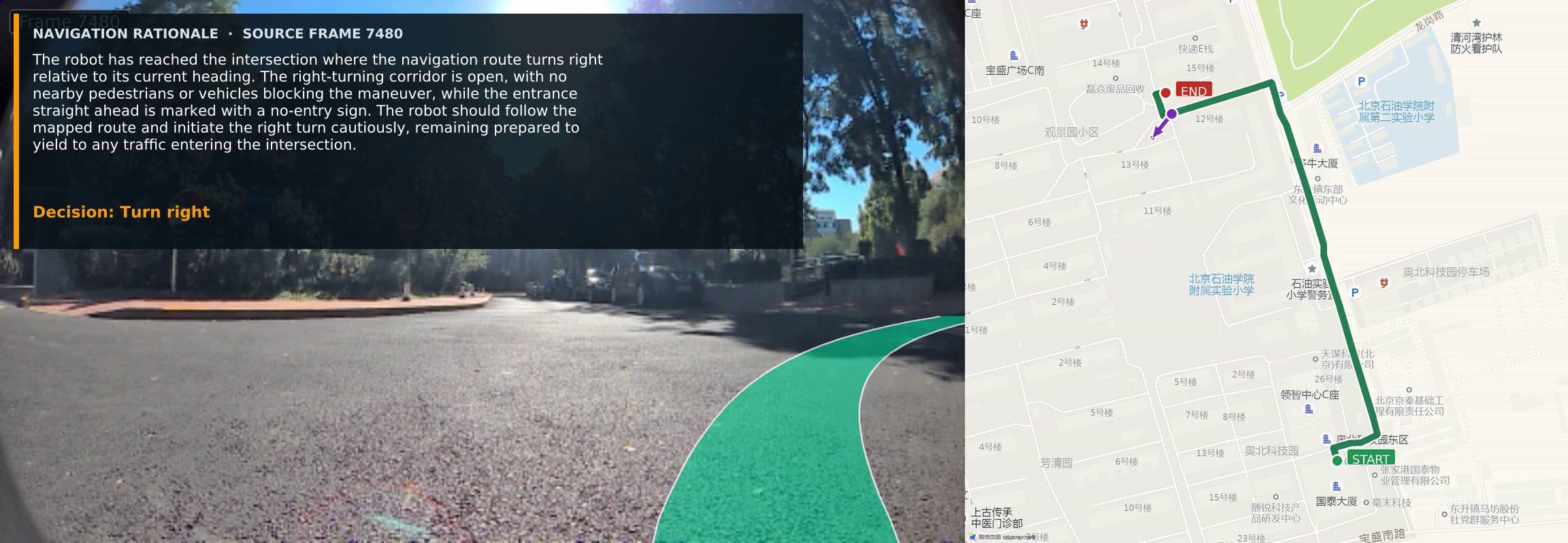}\\[-0.5mm] \includegraphics[width=0.9\linewidth]{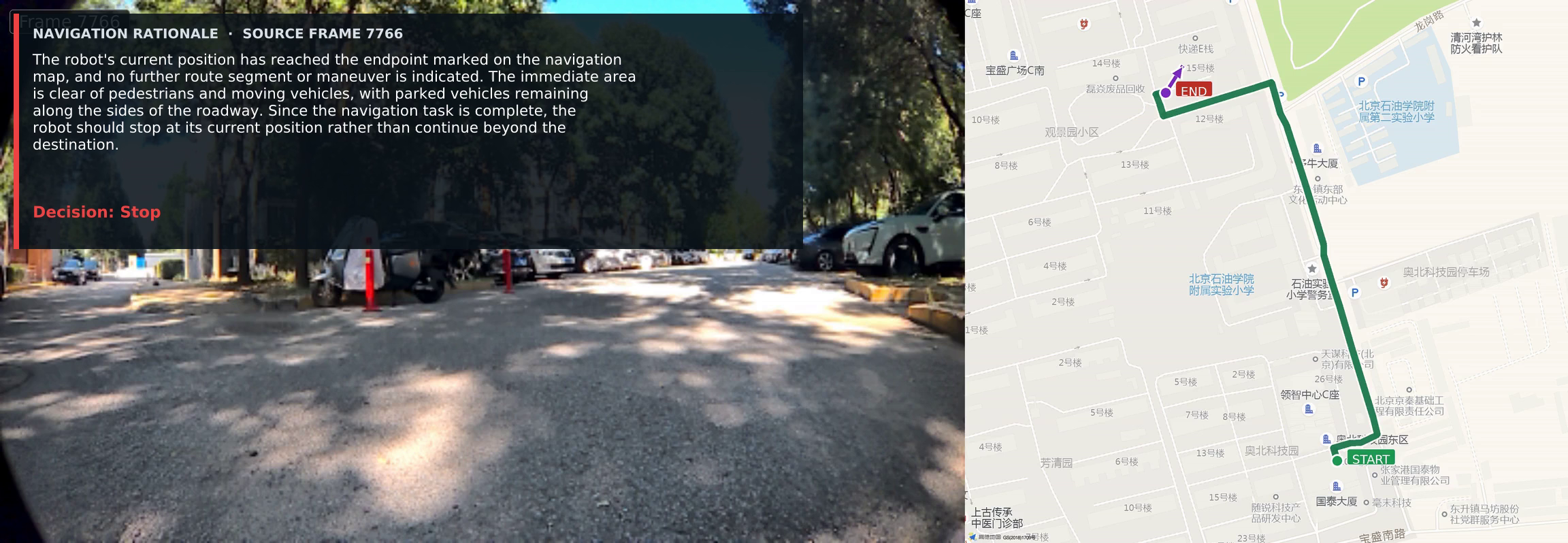}
    \caption{Selected snapshots from an indoor--outdoor navigation rollout. From top to bottom, the robot proceeds through a clear doorway (frame 350), initiates a left turn when the maneuver becomes locally actionable (frame 1202), follows a later right turn at an open intersection (frame 7480), and stops after reaching the route endpoint (frame 7766). In each row, the left panel overlays the predicted local trajectory on the egocentric view and reports the grounded rationale and action; the right panel shows global route progress, current pose, start, and goal.}
    \label{fig:navigation_demo}
\end{figure}

\end{document}